\documentclass[11pt]{article}

\usepackage[final]{acl}

\usepackage[utf8]{inputenc} 
\usepackage[T1]{fontenc}    
\usepackage{hyperref}       
\usepackage{url}            
\usepackage{booktabs}       
\usepackage{amsfonts}       
\usepackage{nicefrac}       
\usepackage{microtype}      
\usepackage{xcolor}         
\usepackage{soul}           

\usepackage{subcaption}
\usepackage{float}
\usepackage{multicol}
\usepackage{amssymb}
\usepackage{amsmath}
\usepackage{graphicx}
\usepackage{threeparttable}
\usepackage{multirow}
\usepackage{makecell}
\usepackage{placeins}
\usepackage{needspace}
\usepackage{adjustbox}
\usepackage{caption}                  
\usepackage{amsthm}

\usepackage{times}
\usepackage{latexsym}

\title{Do Spoken Language Models Hear Speech as They Read Text? \\Bridging Structural Gaps Between Speech and Text}

\author{
 \textbf{Hyeonyu Kim\textsuperscript{1}},
 \textbf{Hwayeon Kim\textsuperscript{1}},
 \textbf{Youngwon Choi\textsuperscript{1}},
 \textbf{Myeongkyun Cho\textsuperscript{1,2}},
 \textbf{Huu-Kim Nguyen\textsuperscript{3}},
\\
 \textsuperscript{1}Maum AI Inc.,
 \textsuperscript{2}KAIST,
 \textsuperscript{3}Atmanity Inc.
\\
 \small{
   \texttt{
   \href{mailto:hykim@maum.ai}{hykim@maum.ai},
   \href{mailto:khy0908@maum.ai}{khy0908@maum.ai},
   \href{mailto:youngwonchoi@maum.ai}{youngwonchoi@maum.ai},
   \href{mailto:mkcho@maum.ai}{mkcho@maum.ai},
   \href{mailto:huukim136@gmail.com}{huukim136@gmail.com}
   }
 }
\\
 \small{
   \textbf{Correspondence:} \href{mailto:hykim@maum.ai}{hykim@maum.ai}
 }
}

\begin{document}
\maketitle

\begin{abstract}

Spoken Language Models (SLMs) generate textual responses directly from speech, offering an alternative to cascaded systems.
Despite recent advances, existing SLMs still exhibit weaker instruction-following behavior and limited generalization across diverse tasks compared to text-based language models.
Our analysis shows that speech and text representations in current SLMs remain weakly aligned despite strong downstream performance, indicating that structural differences between continuous, temporally varying speech and discrete text remain insufficiently addressed.
To address this, we propose a simple framework that decouples length mismatch from semantic alignment and encourages closer correspondence between speech and text representations.
Experiments across multiple benchmarks demonstrate competitive performance against strong baselines, underscoring the importance of explicitly addressing structural differences between speech and text in SLM training.
Our code is publicly available at \url{https://github.com/jaykim9870/Do_SLMs_Hear_Speech_as_They_Read_Text}.

\end{abstract}
    
\vspace{-1em}

\section{Introduction}
\label{sec:intro}

Spoken Language Models (SLMs) have attracted significant attention as a paradigm for enabling more general-purpose interaction with speech.
SLMs broadly encompass pure speech LMs, speech+text LMs, and speech-aware text LMs, which differ in how speech and text are represented and modeled~\cite{arora2025landscape}.
In this work, we focus specifically on \emph{speech-aware text LMs}, which combine a text LLM with a speech encoder to generate textual responses from speech and natural-language instructions, and refer to this class as SLMs throughout the paper.
By processing speech directly rather than relying on an explicit ASR--LLM cascade, these models can mitigate error propagation and preserve speech-specific information such as acoustic and paralinguistic cues~\cite{fathullah2023audiochatllama, wang2024blsp-emo}.

Prior studies \cite{tang2023salmonn, wang2023blsp, yu2024self, pan2023cosmic} have shown that when SLMs are trained on automatic speech recognition (ASR) data to predict the transcription, they tend to focus only on speech content while ignoring textual instructions, a behavior referred to as \emph{speech anchor bias} \cite{yu2024self} or \emph{task overfitting} \cite{tang2023salmonn}.
To mitigate this issue, prior works \cite{fathullah2023audiochatllama, yu2024self, kang2024frozen, lu2025desta2} leverage diverse instructions by generating responses from text descriptions of speech and training SLMs to reproduce the same behaviors directly from speech.
This \emph{behavior alignment} strategy improves instruction-following while allowing SLMs to capture paralinguistic information, even without updating the LLM weights \cite{kang2024frozen, lu2025desta2}.

However, these approaches primarily enforce alignment at the \emph{behavioral} level by encouraging the same responses, while leaving the alignment of internal speech representations implicit.
Recent studies have shown that current SLMs still exhibit notable limitations, including difficulties with instruction-following on Speech-IFEval \cite{lu2025speech}, limited generalization of diverse tasks on Dynamic-SUPERB \cite{huang2024dynamic, huang2024dynamic2}, and a consistent performance gap relative to text-only models on SpeechR \cite{yang2025speechr}.
These observations raise a fundamental question: \emph{Do SLMs Hear Speech as They Read Text?}

We are motivated by this question and analyze the internal representations of speech and text in current SLMs.
As detailed in Section~\ref{sec:method:analysis}, we observe that even when SLMs perform well on downstream tasks, their internal representations of speech remain weakly aligned with text.
This suggests that, despite the strong semantic correspondence between speech and its transcription, current SLMs map speech into representations that remain structurally different from text embeddings.

We argue that structural differences between speech and text are a key factor underlying this discrepancy.
Unlike text, speech is a continuous and time-varying signal, which leads to longer and structurally distinct representations compared to text embeddings.
Prior studies \cite{zhang2023tuning, wang2023slm, tang2023salmonn} have recognized these structural differences and have largely focused on reducing the length of mapped speech features to narrow this gap.

In parallel, other works have explored explicitly treating text embeddings as alignment targets, 
by measuring an L2 distance using only a subset of mapped speech tokens \cite{held2024distilling} or employing the Wasserstein distance \cite{zufle2024contrastive}.  
Alternatively, another line of work \cite{wang2024blsp-kd, deng2024wav2prompt} explicitly aligns sequence lengths using a CIF mechanism and applies an internal alignment loss. However, this design requires the modality adapter to perform length matching and semantic alignment simultaneously.

In this work, we propose a simple framework that explicitly addresses structural differences between speech and text and encourages closer correspondence between the two representations.
Specifically, during training, we dynamically match the length of mapped speech features to text embeddings, thereby decoupling length matching from semantic alignment.
Building on this design, we further incorporate a \emph{token-level internal alignment} alongside \emph{behavior alignment} to encourage more consistent correspondence between speech and text representations.
Experimental results across multiple benchmarks demonstrate that our approach achieves competitive performance against strong baselines, highlighting the importance of explicitly addressing structural differences between speech and text in SLM training.

In summary, our contributions are threefold:

\begin{itemize}
    \item We analyze the internal representations of speech and text in current SLMs and show that they remain weakly aligned, even when models perform well on downstream tasks.
    \item We propose a simple framework that decouples length mismatch from semantic alignment and incorporates \emph{token-level internal alignment}, encouraging closer correspondence between speech and text features.
    \item We demonstrate competitive performance across multiple benchmarks, including comparisons with strong closed-source models, highlighting the importance of explicitly addressing structural differences between speech and text in SLM training.
\end{itemize}
\vspace{-1em}

\section{Related Work}\label{sec:related}

Researchers have explored SLMs with a variety of input/output modality setups and training methods \cite{arora2025landscape}. 
Here, we concentrate on the line of research that generates textual responses from speech inputs.

Early SLM research often involve single-task training such as automatic speech recognition (ASR) or automatic speech translation (AST) \cite{lakhotia2021generative, kharitonov2021text, chang2022speechprompt, wu2023decoder, zhang2023tuning, chen2024salm}. More recent studies have shown that involving multiple speech tasks can enhance SLMs with a broader understanding of spoken language \cite{tang2023salmonn, chu2024qwen2, deshmukh2023pengi}. These architectures integrate a pre-trained speech encoder and a large language model with a modality adapter, and it is common to freeze the pre-trained model \cite{fathullah2023audiochatllama, wang2023blsp, kang2024frozen} or to apply lightweight LoRA-based fine-tuning \cite{gong2023joint, wang2024blsp-emo} to preserve the rich knowledge in each component.

\begin{figure*}[t]
  \centering
  \begin{adjustbox}{width=0.9\linewidth} 
    \begin{minipage}{\linewidth}
      \centering
      \begin{subfigure}[t]{0.24\linewidth}
        \includegraphics[width=\linewidth]{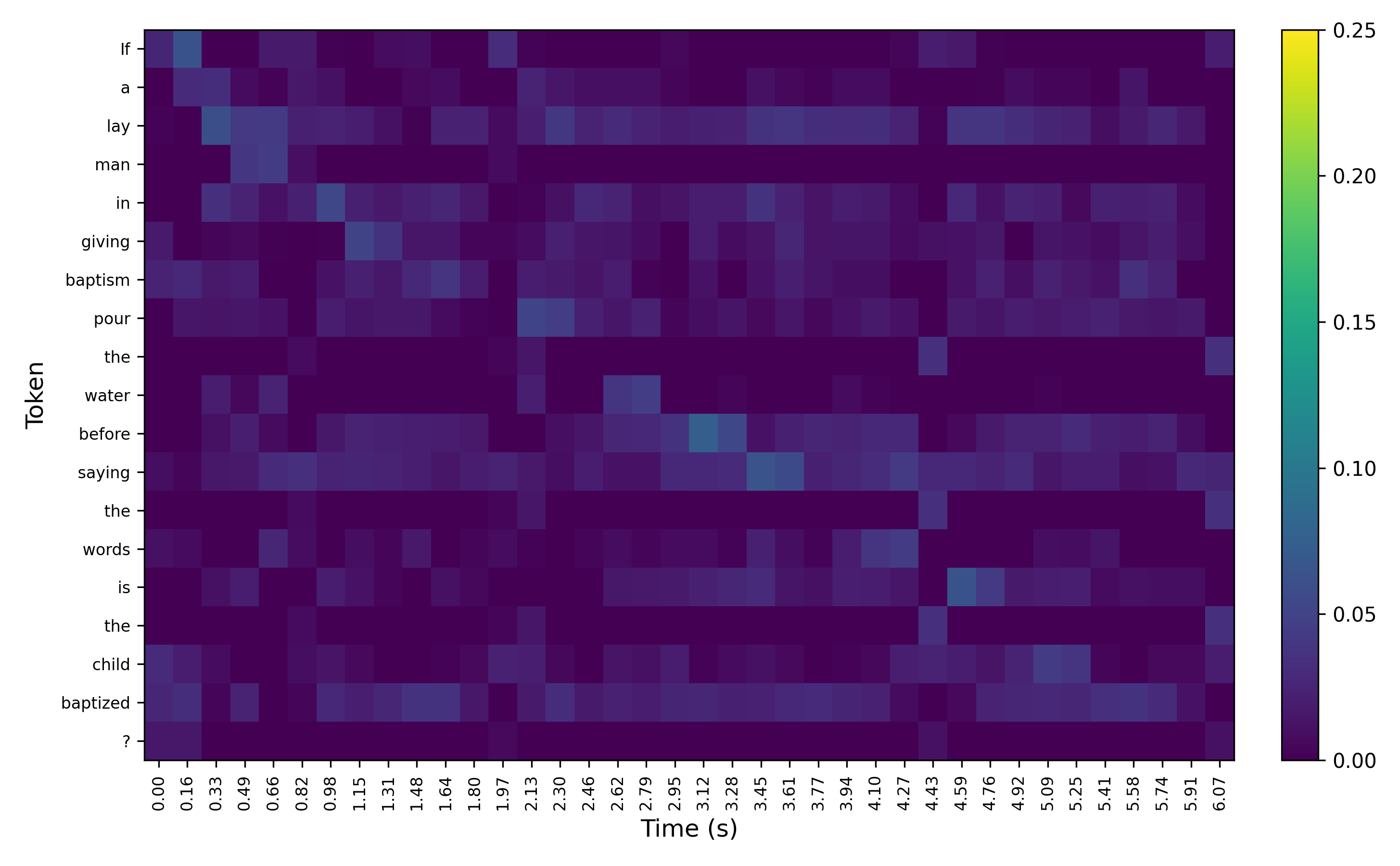}
      \end{subfigure}\hfill
      \begin{subfigure}[t]{0.24\linewidth}
        \includegraphics[width=\linewidth]{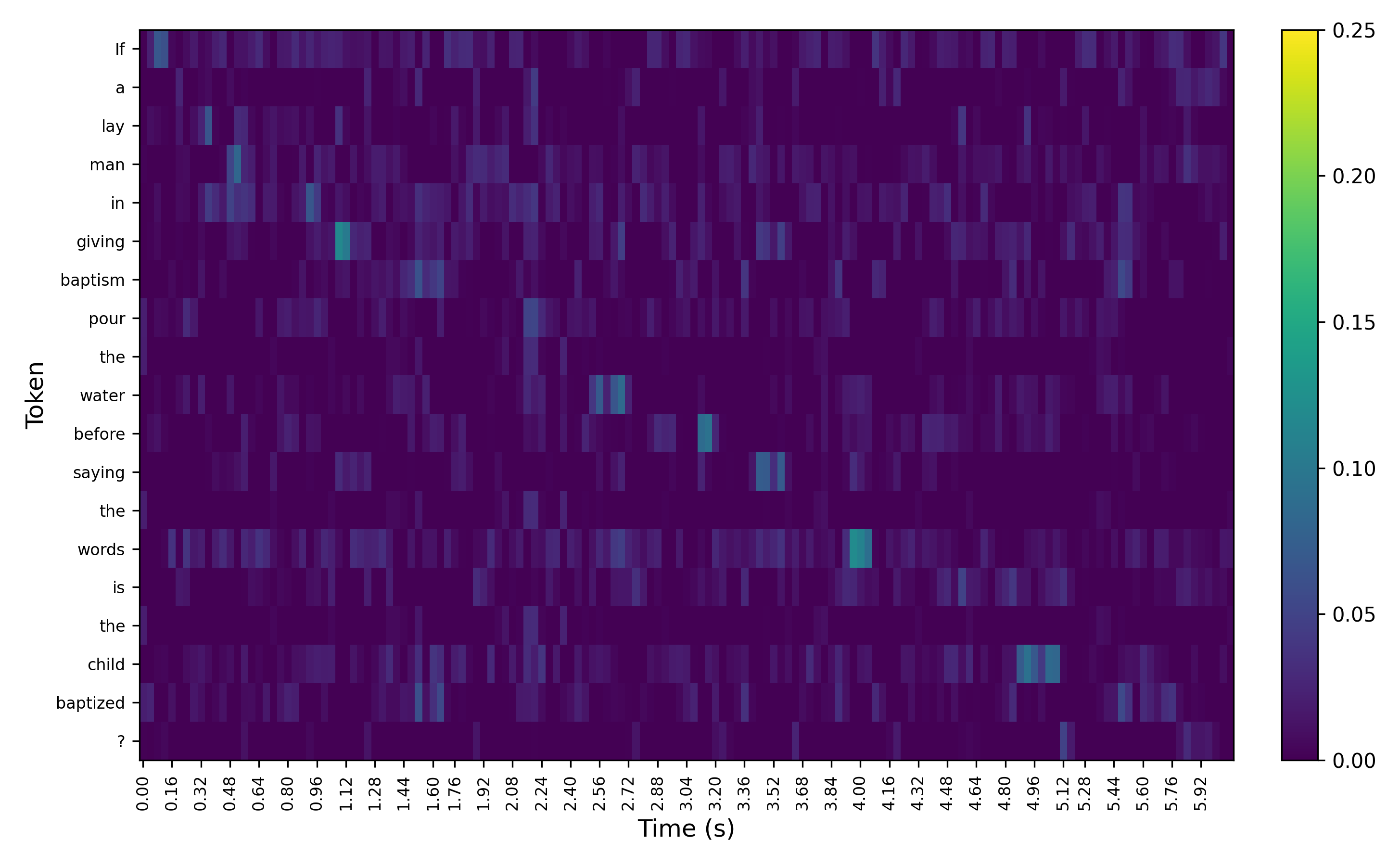}
      \end{subfigure}\hfill
      \begin{subfigure}[t]{0.24\linewidth}
        \includegraphics[width=\linewidth]{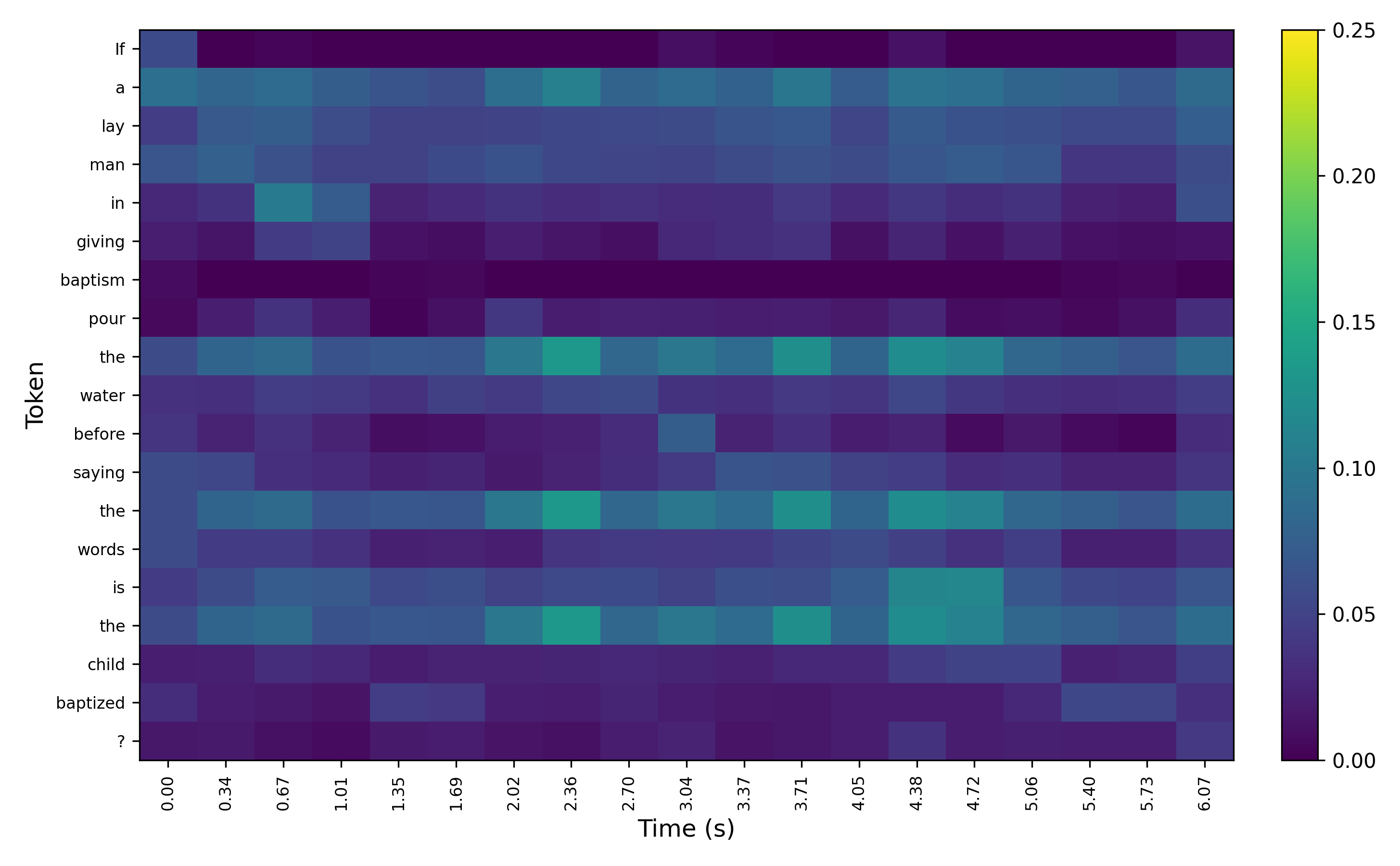}
      \end{subfigure}
      \begin{subfigure}[t]{0.24\linewidth}
        \includegraphics[width=\linewidth]{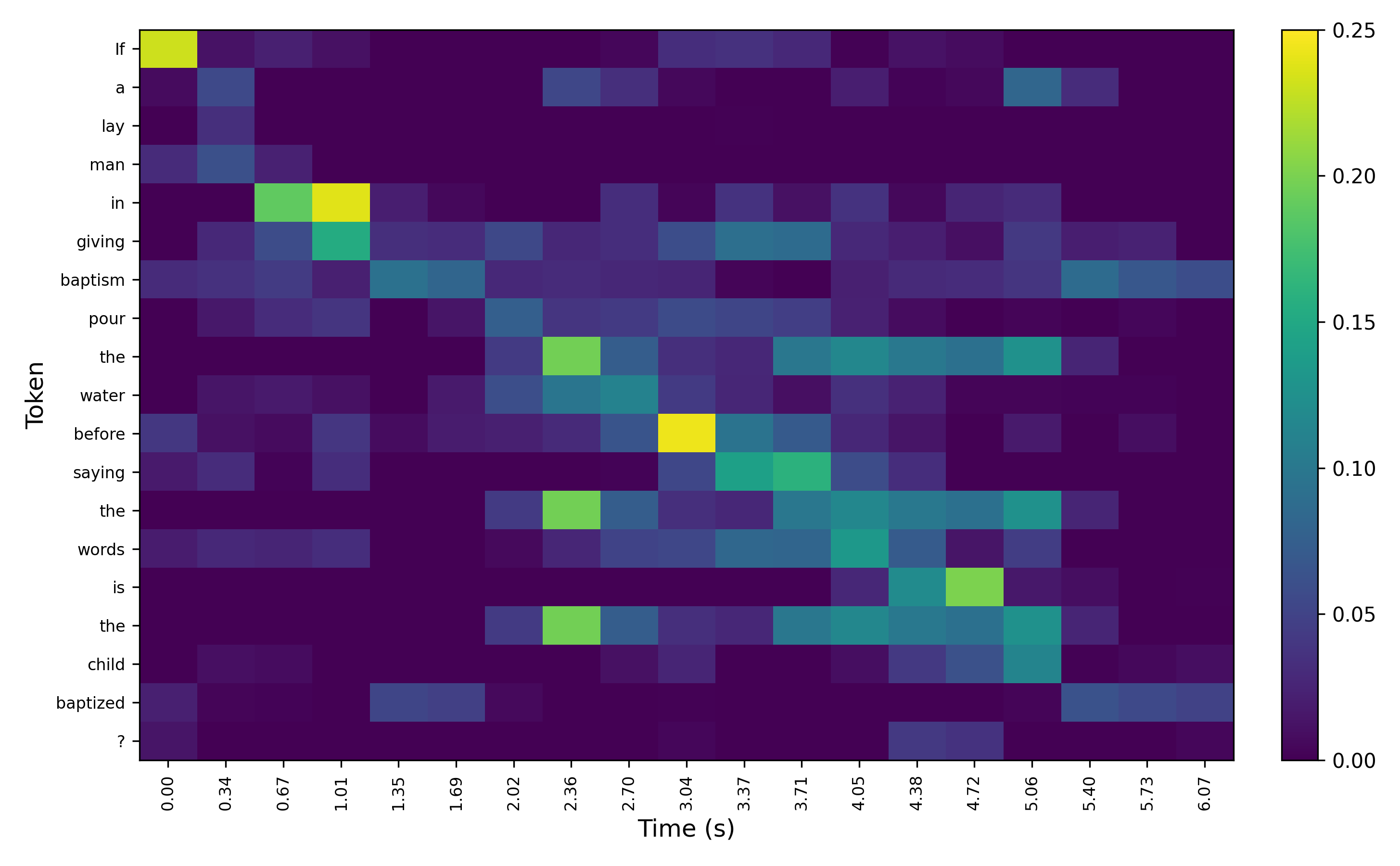}
      \end{subfigure}

      \vspace{0.5em}

      \begin{subfigure}[t]{0.24\linewidth}
        \includegraphics[width=\linewidth]{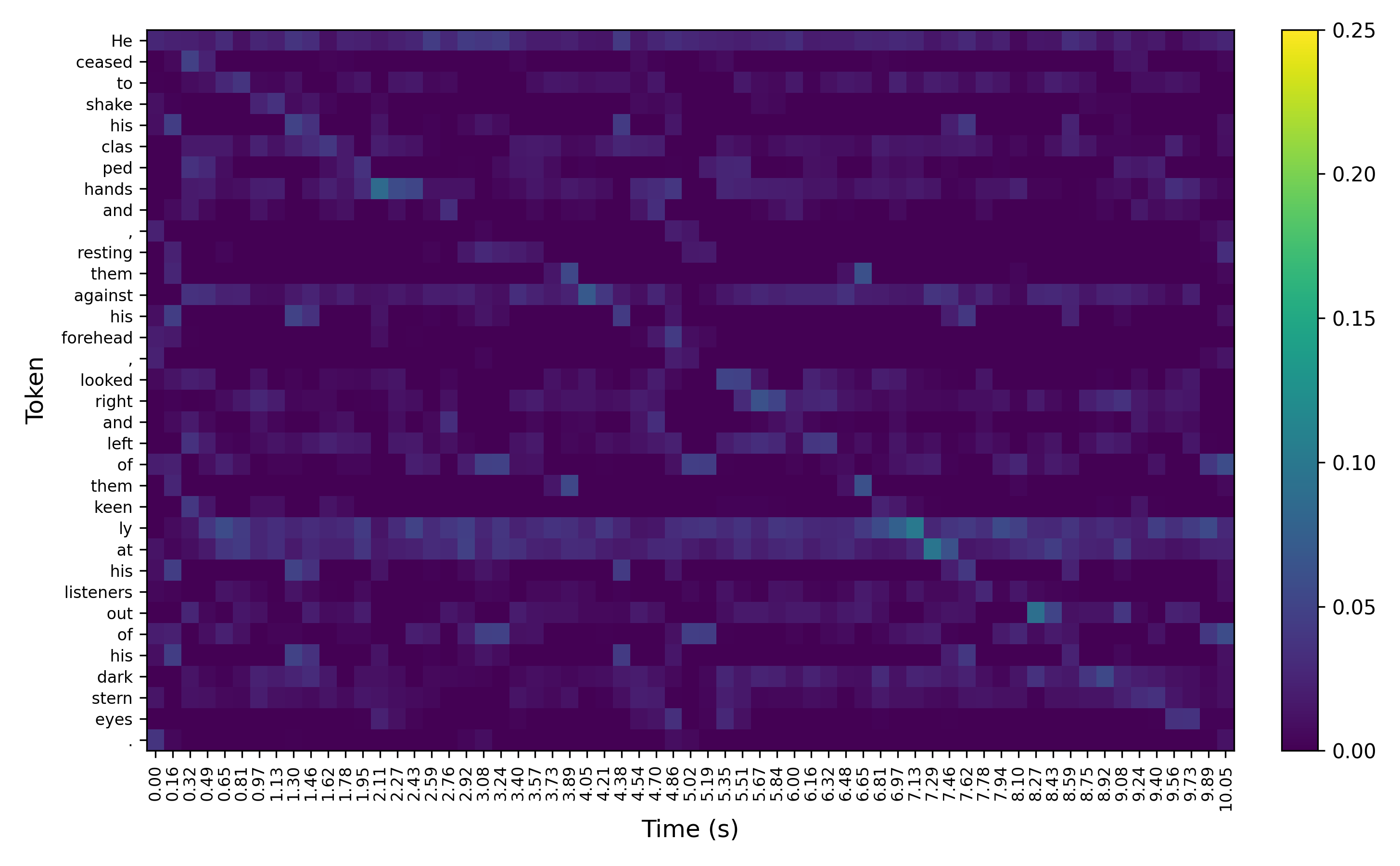}\subcaption{BLSP-emo}
      \end{subfigure}\hfill
      \begin{subfigure}[t]{0.24\linewidth}
        \includegraphics[width=\linewidth]{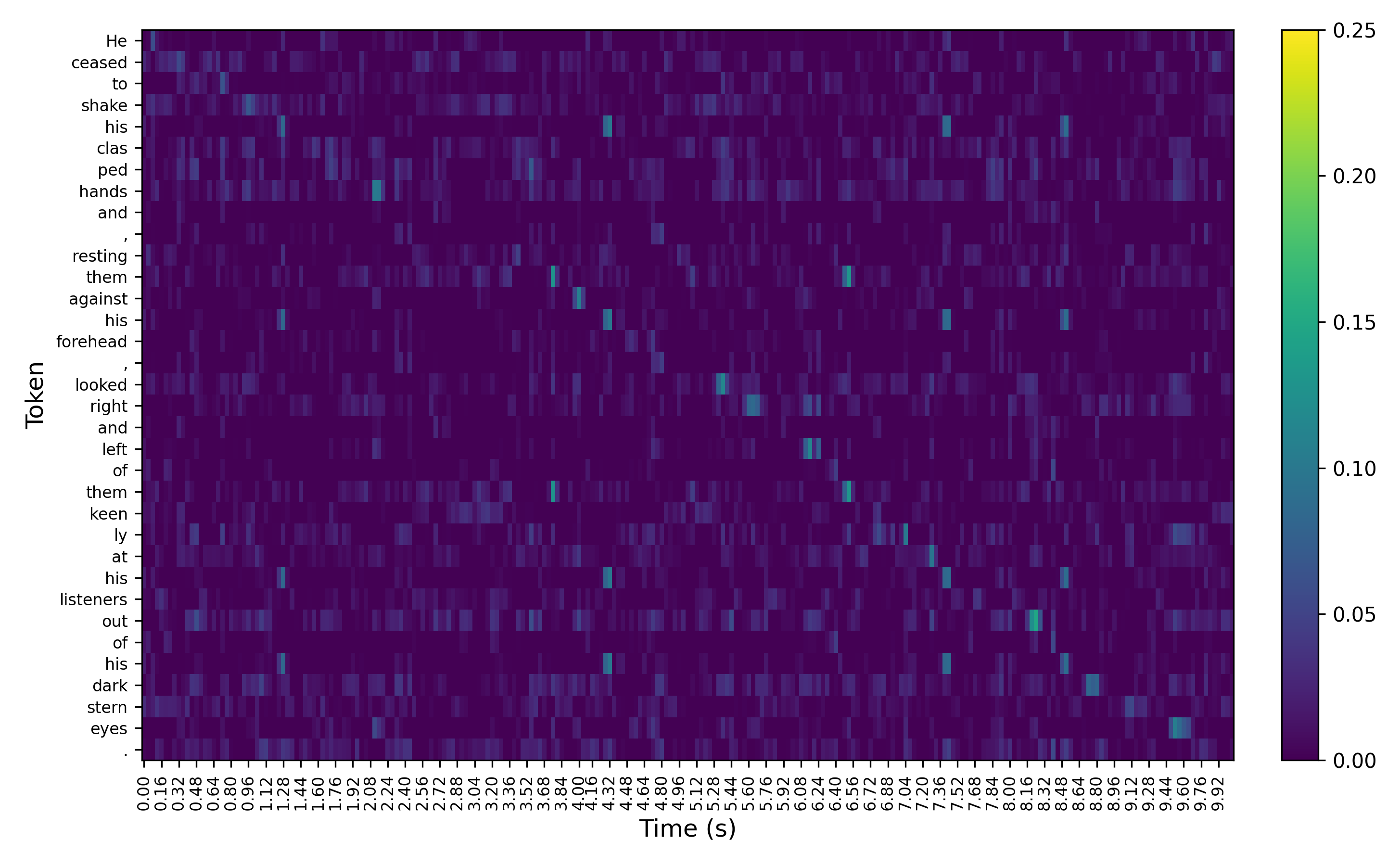}\subcaption{Qwen2-Audio-Instruct}
      \end{subfigure}\hfill
      \begin{subfigure}[t]{0.24\linewidth}
        \includegraphics[width=\linewidth]{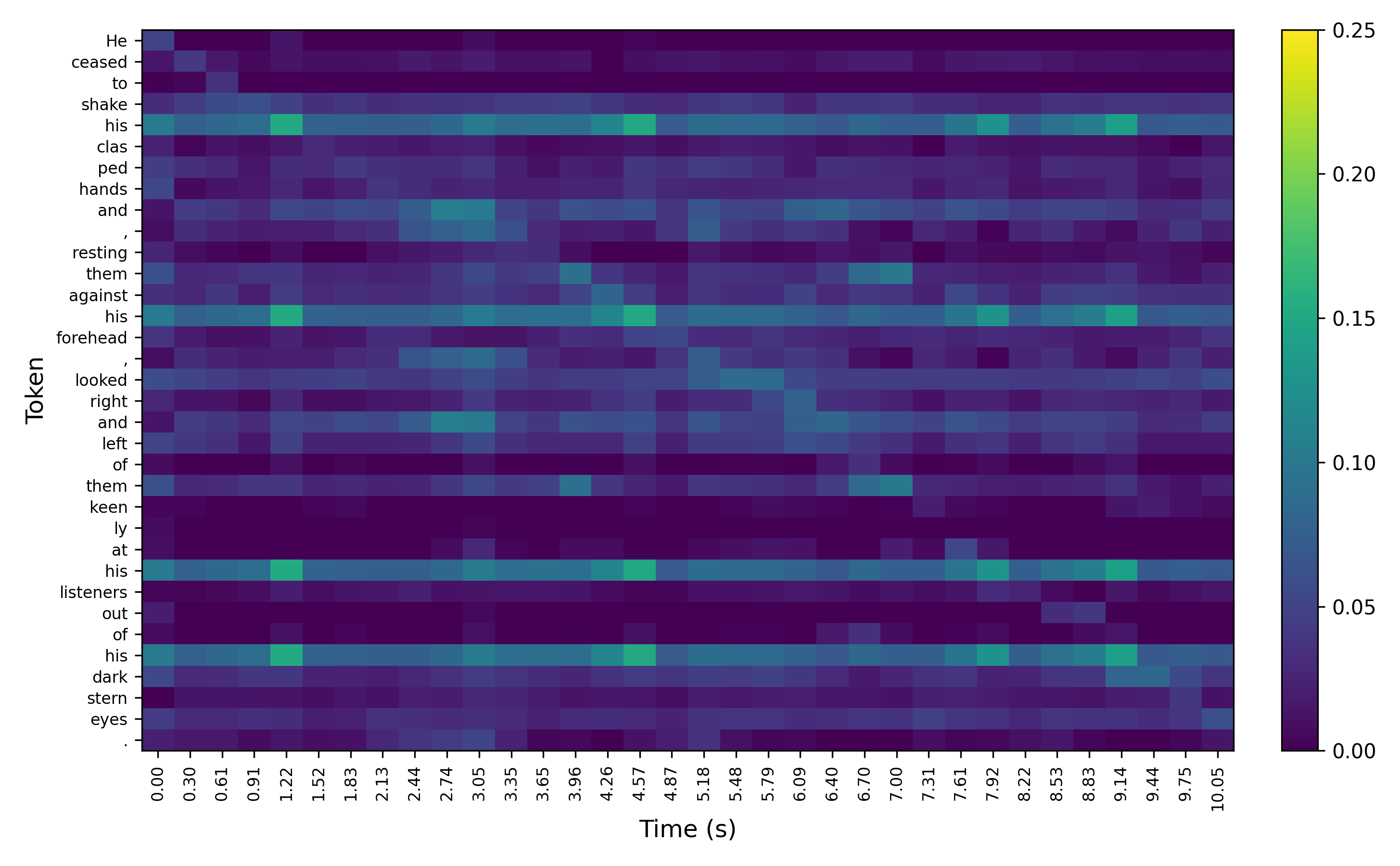}\subcaption{Ours(Cos)}
      \end{subfigure}
      \begin{subfigure}[t]{0.24\linewidth}
        \includegraphics[width=\linewidth]{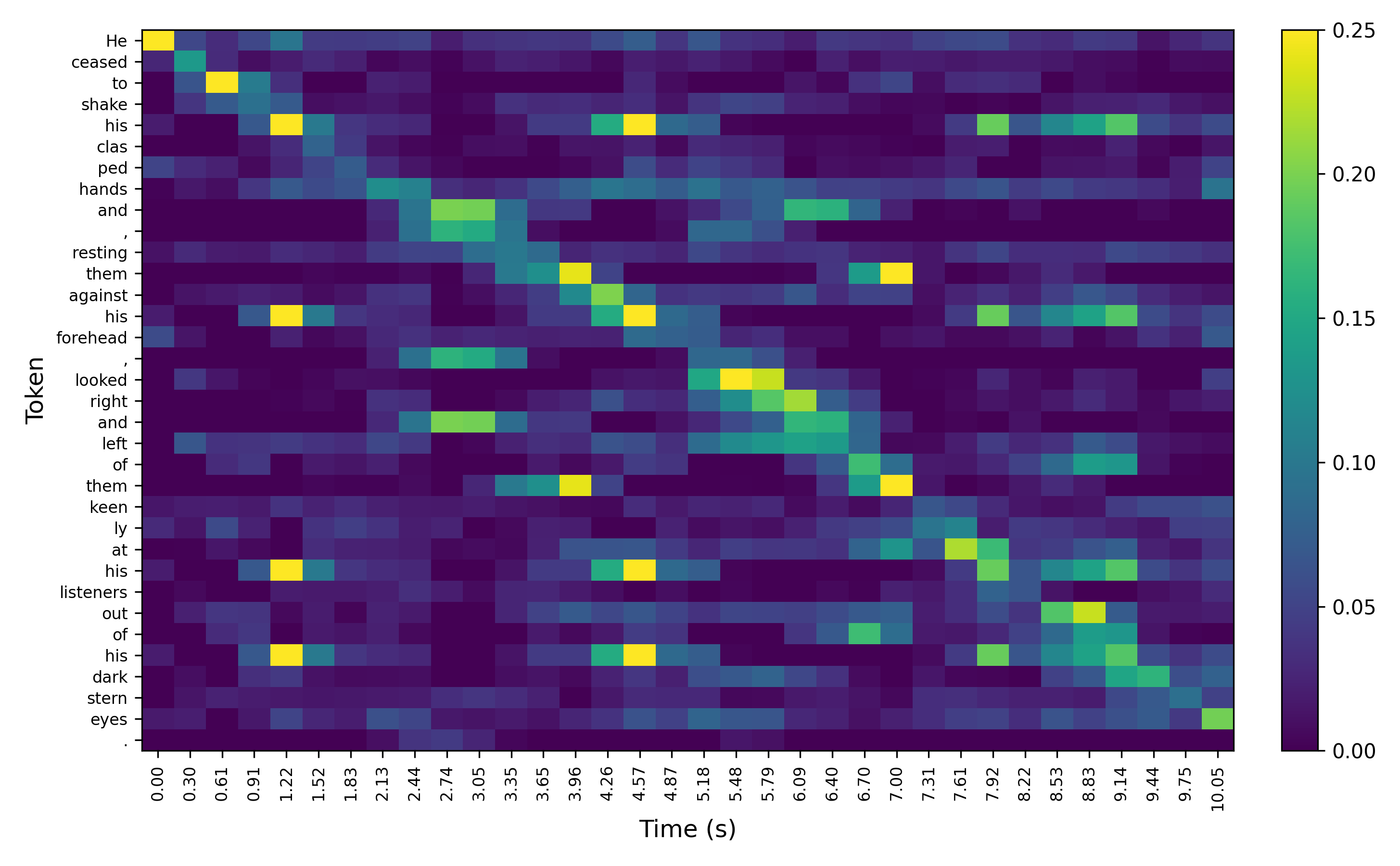}\subcaption{Ours(InfoNCE)}
      \end{subfigure}
    \end{minipage}
  \end{adjustbox}

  \caption{Token-wise similarity maps between $Z_s$ and $Z_t$ on LibriTTS \cite{zen2019libritts} test-clean for BLSP-emo \cite{wang2024blsp-emo}, Qwen2-Audio-Instruct \cite{chu2024qwen2}, and our models. Existing SLMs show weak diagonal patterns, whereas our model exhibits a clear diagonal trend by addressing structural difference between speech and text. Non-diagonal activations correspond to identical text tokens appearing at different positions.}
  \label{fig:speech_text_similarity}
\end{figure*}

A central challenge when stitching an LLM with a speech encoder is the mismatch between the encoded speech representation and the LLM’s input space. This discordance arises not only from differences in semantic representations between speech and text, but also from the substantially longer sequence lengths \cite{wang2024speech}. To address this, many approaches introduce a modality adapter that downsamples speech features \cite{fathullah2023audiochatllama, wang2023slm, ma2024embarrassingly, li2024whisma}, although such temporal compression may not explicitly preserve higher-level linguistic structure. Some studies have proposed techniques to match the length of speech representations to that of the text transcription \cite{wu2023decoder, deng2024wav2prompt, ma2025legoslm, wang2024blsp-kd}. 

In addition to temporal mismatches, ensuring robust instruction-following capabilities in SLM remains a critical challenge. Studies have shown that SLMs trained solely on ASR objectives overlook textual prompts and focus exclusively on speech inputs \cite{tang2023salmonn, wang2023blsp, yu2024self, pan2023cosmic}. To address this, \emph{behavior alignment} frameworks have been introduced, which generate synthetic instruction–response pairs and then train the SLM on those examples \cite{fathullah2023audiochatllama, yu2024self, kang2024frozen, lu2025desta2}. 

\section{Method}\label{sec:method}

\subsection{Do SLMs Hear Speech as They Read Text?}\label{sec:method:analysis}

Given speech input $s$, SLMs encode the input using a speech encoder $Enc(\cdot)$ and project the output into the LLM input space with a modality adapter $\psi(\cdot)$. Previous works have observed that when SLMs are trained on paired data from automatic speech recognition (ASR) $(s, t)$ to predict transcriptions $t$ from speech $s$, as in Equation~\ref{eq:naive_obj}, they focus only on the speech content while ignoring textual instructions, a phenomenon called \emph{speech anchor bias} \cite{yu2024self} or \emph{task overfitting} \cite{tang2023salmonn}.

\vspace{-1em}
\begin{equation}
    \label{eq:naive_obj}
    \mathcal{L}_{\text{asr}} 
    = - \log P\big(t \mid \psi(Enc(s))),
\end{equation}
\vspace{-1.3em}

\begin{table}[t]
\centering
\resizebox{0.8\linewidth}{!}{%
\begin{tabular}{lccc}
\hline
 Model & CKA \\ \hline
 BLSP-emo \cite{wang2024blsp-emo} & 0.3570 \\
 Qwen2-Audio-Instruct \cite{chu2024qwen2} & 0.4113 \\ 
 DiVA \cite{held2024distilling} & 0.3992 \\ 
 DeSTA2 \cite{lu2025developing} & 0.4302 \\ 
 Ours& \textbf{0.6399} \\ 
 \hline
\end{tabular}
}
\caption{Centered Kernel Alignment (CKA) metric between speech and text in the LLM input space.}
\label{tab:cka}
\end{table}

\begin{figure*}[t!]
    \centering
    \includegraphics[trim=0cm 0cm 0cm 0cm , width= 0.9\textwidth, clip]{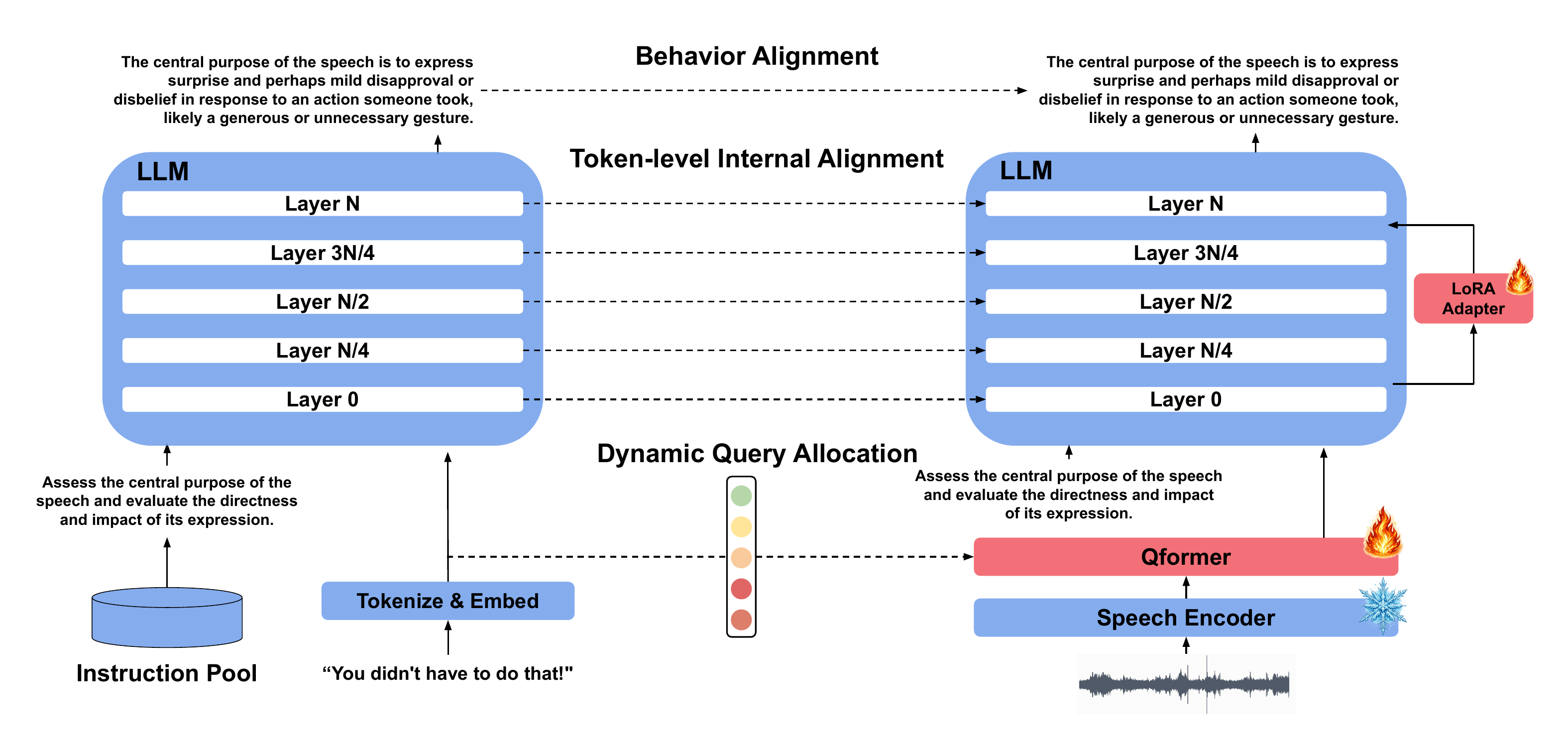}
    \caption{
    Model architecture. Given a speech input, a frozen speech encoder extracts frame-level features, which are mapped into the LLM input space with a windowed Q-former adapter.
    To explicitly resolve length mismatch, the adapter dynamically allocates the number of query tokens to match the target text length during training, decoupling length alignment problem from semantic alignment.
    The model is trained with behavior alignment, together with an token-level internal alignment which encourages fine-grained correspondence between speech and text.
    }
    \label{fig:model_architecture}
\end{figure*}

To mitigate this issue, several works \cite{fathullah2023audiochatllama, yu2024self, kang2024frozen, lu2025desta2} incorporate a diverse instruction set $I$.
Given text descriptions of speech $\tilde{t}$ and an instruction $I_i \in I$, the backbone LLM is used to generate a target response $y_i$, as formalized in Equation~\ref{eq:gen-response}.
The text descriptions $\tilde{t}$ may include not only transcriptions but also additional attributes such as emotion or intent \cite{kang2024frozen, lu2024desta}.

\begin{table*}[t]
\centering
\resizebox{\textwidth}{!}{%
\begin{tabular}{llccc}
\toprule
\textbf{Benchmark} & \textbf{Model} & \multicolumn{3}{c}{} \\
\midrule

\multirow{8}{*}{\textbf{AIR-Bench}} &  & \multicolumn{3}{c}{\textbf{Chat-speech$\uparrow$}} \\ \cmidrule(lr){3-5}
 & SALMONN           & \multicolumn{3}{c}{6.16} \\
 & BLSP              & \multicolumn{3}{c}{6.17} \\
 & DeSTA2            & \multicolumn{3}{c}{7.16} \\
 & Qwen2-Audio       & \multicolumn{3}{c}{7.18} \\
 & Phi-4-Multimodal  & \multicolumn{3}{c}{7.47} \\
 & Gemini-1.5-pro*   & \multicolumn{3}{c}{6.97} \\
 & Gemini-2.0-Flash* & \multicolumn{3}{c}{\textbf{7.92}} \\
 & Ours              & \multicolumn{3}{c}{\underline{7.85}} \\
\midrule

\multirow{6}{*}{\textbf{SpeechR}} &  &
\textbf{Multi-Choice$\uparrow$} &
\makecell{\textbf{Generative-Procedural}\\\textbf{(FC$\uparrow$, LR$\uparrow$, CoH$\uparrow$)}} &
\makecell{\textbf{Generative-Normative}\\\textbf{(FC$\uparrow$, LR$\uparrow$)}} \\
\cmidrule(lr){3-3}\cmidrule(lr){4-4}\cmidrule(lr){5-5}
 & SALMONN              & 34.73 & (12.50, 1.90, 1.33) & (34.75, 3.03) \\
 & Qwen2-Audio          & 12.83 & (9.52, 1.50, 1.00)  & (31.25, 2.82) \\
 & Qwen2-Audio-Instruct & 33.90 & (25.00, 3.50, 2.16) & (38.91, \underline{3.46}) \\
 & Gemini-1.5-Pro*      & \textbf{67.68} & (\textbf{83.04}, \underline{4.49}, \underline{4.47}) & (\textbf{51.92}, \textbf{3.58}) \\
 & Ours                 & \underline{52.91} & (\underline{73.51}, \textbf{4.58}, \textbf{4.79}) & (\underline{40.83}, 3.12) \\
\midrule

\multirow{7}{*}{\textbf{MMSU}} &  &
\textbf{Perception$\uparrow$} & \textbf{Reasoning$\uparrow$} & \textbf{Average$\uparrow$} \\
\cmidrule(lr){3-3}\cmidrule(lr){4-4}\cmidrule(lr){5-5}
 & BLSP                 & 28.36 & 44.77 & 35.96 \\
 & DiVA                 & 33.95 & 65.04 & 48.31 \\
 & Qwen2-Audio-Instruct  & 39.02 & 68.90 & 53.27 \\
 & Gemini-1.5-Pro*      & \textbf{46.10} & \textbf{76.16} & \textbf{60.68} \\
 & Gemini-2.0-Flash*    & \underline{40.83} & 59.18 & 51.03 \\
 & Ours                 & 38.39 & \underline{70.31} & \underline{53.85} \\
\midrule

\multirow{6}{*}{\textbf{Speech-IFEval}} &  &
\textbf{(CEQ$\uparrow$, CW$\uparrow$)} & \textbf{CoT$\uparrow$} & \textbf{Forgetting Rate$\uparrow$} \\
\cmidrule(lr){3-3}\cmidrule(lr){4-4}\cmidrule(lr){5-5}
 & SALMONN              & (37.41, 61.25) & 12.00 & -50.20 \\
 & BLSP-emo             & (66.35, 63.75) & 50.50 & -17.92 \\
 & Qwen2-Audio-Instruct & (41.59, 67.75) & 32.00 & - \\
 & DeSTA2               & (\underline{83.71}, \textbf{92.49}) & \textbf{91.50} & \textbf{-3.57} \\
 & Ours                 & (\textbf{96.14}, \underline{69.50}) & \underline{67.50} & \underline{-12.18} \\
\bottomrule
\end{tabular}
}
\caption{Performances on Air-bench chat, SpeechR, MMSU, and Speech-IFeval benchmarks. Best results are highlighted in \textbf{bold}, and second-best results are \underline{underlined}. 
For SpeechR, FC, LR, and CoH denote final correctness, logical relevance, and coherence, respectively. 
For Speech-IFEval, CEQ, CW, and CoT indicate close-ended question, creative writing, and chain of thought, respectively. 
Models marked with * are closed-source models.}
\label{tab:main}
\end{table*}

\vspace{-1em}
\begin{equation}
\label{eq:gen-response}
y_i \sim P\big(y \mid \tilde{t}, I_i\big)
\end{equation}

Then the SLM is trained to predict $y_i$ directly from speech, as in Equation~\ref{eq:behavior-alignment}. This \emph{behavior alignment} \cite{wang2023blsp} strategy has been shown to improve the instruction-following capability of SLMs, and demonstrates that models can capture paralinguistic information even without updating the LLM weights \cite{kang2024frozen}.

\vspace{-1.5em}
\begin{equation}
\label{eq:behavior-alignment}
\mathcal{L}_{\text{behavior}} 
= - \log P\big(y_i \mid \psi(Enc(s)), I_i\big),
\end{equation}

However, these behavior-alignment approaches mainly encourage the model to produce the same responses from speech inputs, while the alignment between speech and text representations inside the model remains implicit.
Recent studies report that current SLMs still struggle with simple instruction-following \cite{lu2025speech}, and exhibit limited generalization across dynamic tasks \cite{huang2024dynamic, huang2024dynamic2} compared to text-only LLMs. 
These observations motivate a fundamental question: \emph{Do SLMs hear speech as they read text?}

To examine this, we compare mapped speech features $Z_s=\psi(Enc(s))$ in the LLM input space with text embeddings $Z_t$. We first measure their similarity using CKA \cite{kornblith2019similarity, raghu2021vision}, which measures shared subspace structure. Existing SLMs exhibit low CKA scores (Table~\ref{tab:cka}), which indicates limited structural similarity between speech and text representations.

To obtain a more fine-grained perspective, we visualize the token-wise similarity maps between $Z_s$ and $Z_t$.
As shown in Figure~\ref{fig:speech_text_similarity}, existing SLMs exhibit weak or inconsistent diagonal patterns.
This observation implies that the learned cross-modal mapping in current SLMs may yield representations with a different structure from text, rather than naturally aligning with them at the token level.

\subsection{Persistent Structural Differences in SLMs}\label{sec:method:length_align}

In this section, we take a closer look at structural differences between speech and text that may contribute to this discrepancy.
That is, speech consists of continuous and temporally varying signals and the encoded speech $Enc(s) \in \mathbb{R}^{L_s \times d_s}$ is much longer and structurally different from their textual counterparts, which are discrete and symbolic.

Prior approaches reduce the length of mapped speech features using convolutional layers \cite{zhang2023tuning, das2024speechverse}, downsampling \cite{wang2023slm, gong2023listen, kang2024frozen}, or a windowed Q-former \cite{tang2023salmonn, wang2025u, mousavi2025listen}. 
For example, Qwen2-Audio \cite{chu2024qwen2} produces 25 tokens per second of audio, while DeSTA2 \cite{held2024distilling} yields 64 tokens regardless of input length. 
However, such compression can merge multiple subword or phonetic units into a single token, making the representation sensitive to temporal variations such as hesitations, stuttering, elongated vowels, or speaking rate changes.
Recent methods further explore length reduction using CTC posteriors \cite{ma2025legoslm} or residual vector quantization (RVQ) \cite{tseng2025taste}, but they still primarily focus on alleviating length mismatch as a way to address speech-text structural differences.

In parallel, several studies have attempted to improve cross-modal alignment by explicitly treating text embeddings as alignment targets. 
For instance, DiVA \cite{held2024distilling} measures an L2 distance using only a subset of mapped speech features, while \citet{zufle2024contrastive} employ the Wasserstein distance to compare representations with different lengths. 
Alternatively, methods that explicitly align sequence lengths using CIF mechanisms allow KL divergence \cite{wang2024blsp-kd} or mean squared error \cite{deng2024wav2prompt} to be directly applied. 
However, these approaches rely on additional objectives to train the CIF module itself and require the adapter to perform length matching and semantic alignment simultaneously.

\subsection{Our Approach}\label{sec:method:ours}

In the previous sections, we show that existing SLMs process speech differently from text (Section~\ref{sec:method:analysis}), and that structural differences between speech and text can persist (Section~\ref{sec:method:length_align}). 
In this work, we argue that explicitly addressing these structural differences can improve cross-modal alignment between speech and text.
To this end, we introduce a simple framework that decouples length mismatch from semantic alignment and encourages closer correspondence between speech and text representations.
Figure~\ref{fig:model_architecture} illustrates the overall architecture.

Our modality adapter $\psi(\cdot)$ is based on a windowed Q-former \cite{tang2023salmonn}, in which mapped speech features attend to speech representations via cross-attention and their output length is determined by the learned query. 
Unlike prior approaches that rely on a fixed rate of query allocation, we adopt a \emph{dynamic query allocation} strategy to adjust the length of mapped speech features. 
During training, we leverage speech--transcription pairs $(s, t)$ and allocate $L_t$ queries to match the length of the target text embeddings $Z_t \in \mathbb{R}^{L_t \times d}$, explicitly controlling the length of mapped speech features.
At inference time, we employ a lightweight speech rate predictor \cite{yeo2025mms} to estimate the target token length from speech and allocate queries accordingly. 
Details of the speech rate predictor are provided in Appendix~\ref{sec:appendix:implementation}, and its performance is reported in Table~\ref{tab:srp_performance}. 
By decoupling length alignment from semantic alignment, this simple design facilitates learning meaningful semantic correspondences between speech and text.

\begin{table*}[t]
\centering
\renewcommand{\arraystretch}{1.15}
\resizebox{0.7\textwidth}{!}{%
\begin{tabular}{lccccc}
\toprule
 & \textbf{Air-bench} & \textbf{SpeechR} & \textbf{MMSU} & \textbf{Speech-IFeval} & \textbf{Rel. $\Delta$ (\%) $\uparrow$} \\
\midrule
Ours(w/o $\mathcal{L}_{\text{behavior}}$) & 7.59 & 47.33 & 52.45 & -22.94 & -16.20 \\
Ours(w/o $\mathcal{L}_{\text{internal}}$) & 7.24 & 52.00 & 53.14 & -17.68 & -10.66 \\ 
\midrule
Ours(MSE) & 7.66 & 48.46 & 53.63 & -24.51 & -15.59 \\
Ours(InfoNCE) & 7.48 & 48.75 & 53.67 & -23.30 & -15.39 \\ 
\midrule
Ours(Cformer) & 5.07 & 36.13 & 39.97 & -29.60 & -48.71 \\
Ours(FQA) & 7.78 & 52.41 & 53.25 & -18.26 & -9.07 \\
Ours(w/o SRP) & \textbf{7.89} & 51.93 & \textbf{54.24} & -14.64 & -4.37 \\ 
\textcolor{gray}{Ours(w/o SRP, GT)} & \textcolor{gray}{7.85} & \textcolor{gray}{53.05} & \textcolor{gray}{53.69} & \textcolor{gray}{-12.70} & \textcolor{gray}{-1.07} \\ 
\midrule
\textbf{Ours} & 7.85 & \textbf{52.91} & 53.85 & \textbf{-12.18} & \textbf{0.00} \\
\bottomrule
\end{tabular}%
}
\caption{Ablation study across four benchmarks. We report performance on each benchmark and relative changes (\%) with respect to the full model, where higher values indicate better performance.}
\label{tab:ablation_results}
\end{table*}

We adopt \emph{behavior alignment} and train our model to generate the same LLM response $y$ produced from the text descriptions of speech $\tilde{t}$, as defined in Equation~\ref{eq:behavior-alignment}. 
Note that we randomly sample instruction-response pair $(I_i, y_i)$ from instruction set $I$ during training, as illustrated in Figure~\ref{fig:model_architecture}, but for brevity we omit the instruction index below.

\vspace{-1.3em}
\begin{equation}
\label{eq:ce}
\resizebox{0.87\linewidth}{!}{$
\begin{aligned}
\mathcal{L}_{\mathrm{behavior}}
= - \sum_{j=1}^{T}
\log P\big(y_j \mid y_{<j}, \psi(Enc(s)), I\big)
\end{aligned}
$}
\end{equation}

In addition, we apply a fine-grained \emph{token-level internal alignment} loss between speech and text representations.
Specifically, we use a cosine similarity loss, defined in Equation~\ref{eq:cosine}, to encourage alignment between the length-matched speech and text representations at selected token-level hidden states $h_s^{(\ell)}$ and $h_t^{(\ell)}$ of the LLM.
Since dynamic query allocation matches the mapped speech sequence length to the target text length $L_t$ during training, the loss is computed over corresponding token positions.
We compute the token-level internal alignment loss at the input embedding layer, denoted as $\mathcal{L}_{\mathrm{internal}}^{(0)}$, as well as at four evenly spaced hidden layers, $N/4$, $N/2$, $3N/4$, and $N$, where $N$ denotes the number of layers in the LLM.
The losses are obtained by averaging across layers.

\begin{equation}
\label{eq:cosine}
\resizebox{0.8\linewidth}{!}{$
\begin{aligned}
\mathcal{L}_{\mathrm{internal}}^{(\ell)}
=
\frac{1}{L_t}
\sum_{j=1}^{L_t}
\left(
1 -
\frac{
\langle h_{s,j}^{(\ell)},\, h_{t,j}^{(\ell)} \rangle
}{
\| h_{s,j}^{(\ell)} \|_2 \, \| h_{t,j}^{(\ell)} \|_2
}
\right).
\end{aligned}
$}
\end{equation}

\noindent The final training objective is defined as below:

\begin{equation}
\label{eq:final_loss}
\mathcal{L}
= \mathcal{L}_{\mathrm{behavior}}
+ \lambda \, \mathcal{L}_{\mathrm{internal}},
\end{equation}

where $\lambda$ controls the relative strength of the token-level internal alignment objective.

\begin{figure}[t]
    \centering
    \includegraphics[width=0.7\linewidth]{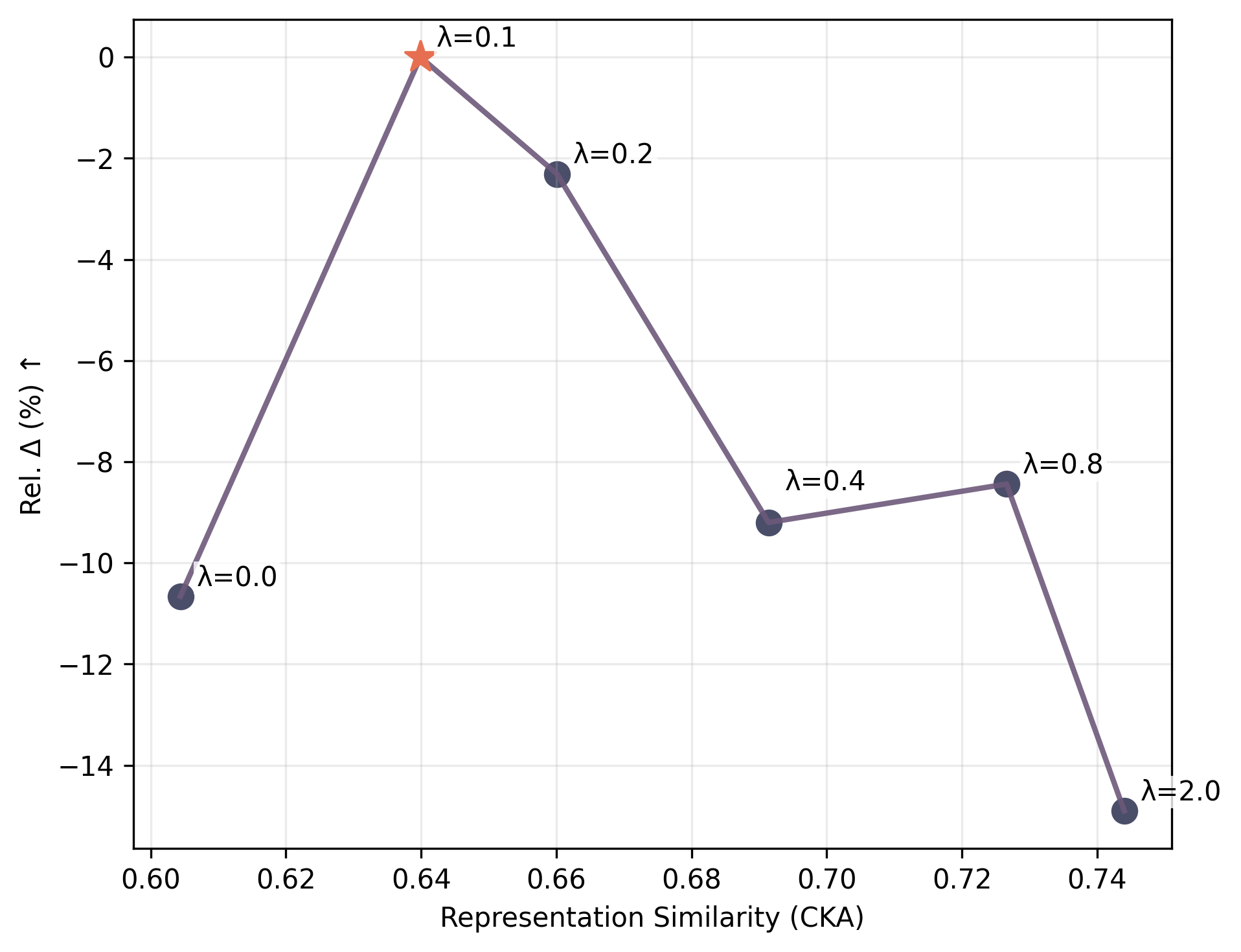}
    \caption{
    Trade-off between representation similarity and task performance across different values of $\lambda$.
    }
    \label{fig:lambda_ablation}
\end{figure}

\section{Experiments} \label{sec:Experiments}

\subsection{Implementation Details}\label{sec:Experiments:implementation}

We use a total of approximately 69{,}000 hours of paired speech--text data, which includes diverse speech-specific attributes. The instruction set $I$ consists of 18 instructions, constructed based on prior work~\cite{yu2024self}. 
For the LLM backbone, we employ Qwen2.5-7B-Instruct~\cite{qwen2.5} and use Whisper-large-v3~\cite{radford2022whisper} as the speech encoder. 
Further details on the datasets and training are provided in Appendix~\ref{sec:appendix:implementation}.

\begin{figure}[t]
    \centering
    \includegraphics[width=0.77\linewidth]{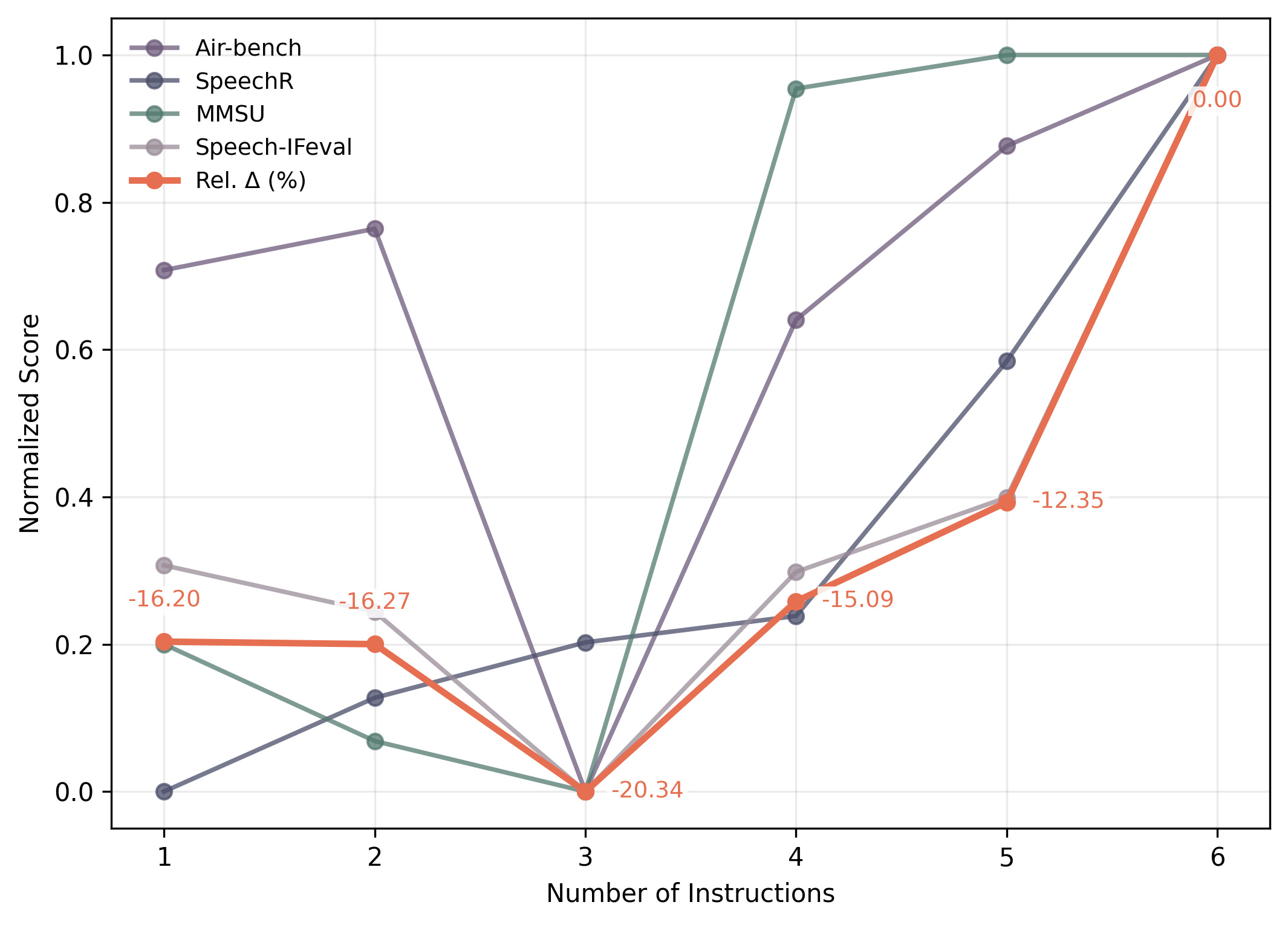}
    \caption{
    Effect of the number of instructions on model performance.
    }
    \label{fig:inst_ablation}
\end{figure}

\begin{table*}[t]
\centering
\small
\resizebox{\textwidth}{!}{%
\begin{tabular}{lcccccc}
\toprule
Model &
\multicolumn{2}{c}{AIR-Bench Foundation} &
AIR-Bench &
SpeechR$\uparrow$ &
MMSU$\uparrow$ &
CKA\\
\cmidrule(lr){2-3}
& Linguistic $\uparrow$ & Speech-specific $\uparrow$
& Chat-speech $\uparrow$ &  & & \\
\midrule
Ours (text input)
& \textbf{70.35} & 44.11 & \textbf{8.43} & \textbf{58.85} & 34.43 & - \\
Ours (w/o $\mathcal{L}_{\mathrm{behavior}}$)
& 53.09 & 47.90 & 7.59 & 47.33 & 52.45 & 0.6436\\
Ours (InfoNCE)
& 52.94 & 47.89 & 7.48 & 48.75 & 53.67 & 0.7113\\
Ours
& 53.33 & \textbf{50.63} & 7.85 & 52.91 & \textbf{53.85} & 0.6399 \\
\bottomrule
\end{tabular}
}
\caption{
Analysis of linguistic and speech-specific performance.
The text-input variant performs strongly on tasks primarily requiring linguistic content, but underperforms on tasks that rely more heavily on speech-specific information.
}
\label{tab:linguistic_speech_specific}
\end{table*}

\subsection{Benchmarks}\label{sec:Experiments:benchmark}

\textbf{AIR-Bench (Chat-speech)}
\cite{yang2024air} contains open-ended question--answer pairs designed to evaluate instruction-following and generative interaction capabilities from audio, and is presented as the first generative benchmark for SLMs.

\noindent \textbf{SpeechR}
\cite{yang2025speechr} is designed to evaluate speech-based reasoning capabilities of SLMs and is constructed using synthetic speech data. 
SpeechR evaluates three types of reasoning, factual retrieval, procedural inference, and normative judgment, and consists of three subsets of multiple-choice, generative, and acoustic-feature formats.

\noindent \textbf{MMSU}
\cite{wang2025mmsu} is a comprehensive benchmark that emphasizes the understanding of speech with diverse acoustic and paralinguistic signals, and is mostly built on real-world speech data. 
It contains 5,000 expert-annotated multiple-choice questions spanning 47 tasks that cover both perception and reasoning.

\noindent \textbf{Speech-IFEval}
\citep{lu2025speech} is designed to evaluate the instruction-following capability of SLMs. 
It disentangles instruction-following from speech perception and introduces instruction constraints that are independent of the speech content, enabling a focused assessment of whether models correctly follow textual instructions.

\subsection{Main Results}\label{sec:Experiments:main_results}

Table~\ref{tab:cka} and Figure~\ref{fig:speech_text_similarity} show that our model exhibits stronger structural similarity between two representations, as evidenced by higher CKA scores and more consistent token-level similarity patterns.

Table~\ref{tab:main} summarizes performance across four benchmarks. Our model achieves competitive results on most benchmarks, including comparisons with strong closed-source models.

SALMONN \cite{tang2023salmonn} employs a window-level Q-former as a modality adapter and introduces behavior alignment to mitigate task overfitting. 
Our model outperforms SALMONN on most benchmarks, with a particularly large improvement in forgetting rate on Speech-IFEval.

Qwen2-Audio-Instruct \cite{chu2024qwen2} is trained via a multi-stage pipeline with large-scale pre-training, supervised fine-tuning, and preference optimization. 
Despite using substantially less training data, our model achieves competitive performance and surpasses it on several subsets on SpeechR and Speech-IFEval.

DeSTA2 \cite{lu2025developing} augments textual descriptions of speech using auxiliary models for better training, and incorporates ASR output along with the audio for inference. 
While DeSTA2 benefits from transcription access, our model achieves stronger performance on AIR-Bench and remains competitive on Speech-IFEval.

Finally, our model demonstrates competitive performance against strong closed-source baselines such as Gemini-1.5-Pro and Gemini-2.0-Flash across multiple benchmarks.
Overall, these results suggest that explicitly addressing structural differences between speech and text can improve performance across diverse tasks.

\subsection{Ablation Study}\label{sec:Experiments:ablation}

In this section, we conduct controlled ablations on key components of our model to analyze their contributions and training dynamics. 
The results are summarized in Table~\ref{tab:ablation_results}, where we report multiple-choice accuracy on SpeechR, average scores on MMSU, and the forgetting rate on Speech-IFEval.

\begin{figure*}[t!]
    \centering
    \includegraphics[trim=0cm 0cm 0cm 0cm , width= \textwidth, clip]{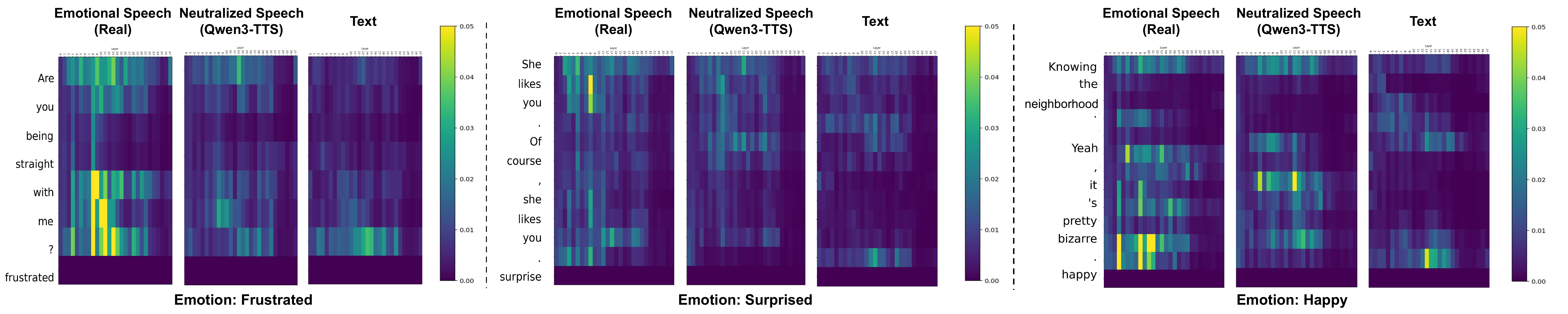}
    \caption{
        Attention analysis on IEMOCAP across real speech, neutralized cloned speech generated by Qwen3-TTS \cite{hu2026qwen3}, and text.
        Brighter colors indicate higher attention weights from the input representations to the corresponding emotion label.
        Our model shows clear activations around regions associated with emotional expression in real speech.
    }
    \label{fig:analysis_iemocap}
\end{figure*}

\subsubsection{Effects of Alignment Objectives and Instruction Diversity}\label{sec:Experiments:ablation:loss}

We first train variants of our model by removing each term in Equation~\ref{eq:final_loss}, and results are shown in the top rows of Table~\ref{tab:ablation_results}. 
Both settings lead to over 10\% performance degradation, with a larger drop observed when excluding the behavior alignment.

We further analyze the effect of $\mathcal{L}_{\mathrm{internal}}$ by varying $\lambda$. 
As shown in Figure~\ref{fig:lambda_ablation}, increasing $\lambda$ consistently improves CKA, indicating stronger representational similarity between speech and text.
However, excessively large $\lambda$ values lead to performance degradation, suggesting that internal alignment is beneficial only at an appropriate strength, as overly strong alignment can compromise overall performance, potentially by reducing the preservation of speech-specific information.

Finally, we study the effect of instruction diversity by progressively expanding the instruction set.
Among the 18 instructions spanning six categories, we add one category at a time, with three instructions per category.
The categories are introduced in order of increasing semantic complexity: Speech Recognition, Content Repetition, Continuation, Keyword Extraction, Intent Recognition, and Sentiment Analysis.
As shown in Figure~\ref{fig:inst_ablation}, performance improves as instruction diversity increases, with noticeable gains when more complex instructions such as Keyword Extraction are introduced.
These results suggest that adding instructions requiring more structured understanding can further improve performance, and that the best results in our setting are obtained by combining behavior alignment with token-level internal alignment.
Detailed results are provided in Table~\ref{tab:inst_ablation}.

\subsubsection{Effect on Different Types of Token-level Internal Alignment Losses}\label{sec:Experiments:ablation:internal}

We explore different types of token-level internal alignment losses and analyze their impact. 
In addition to Equation~\ref{eq:cosine}, we consider an MSE loss that enforces alignment in Euclidean space and an InfoNCE loss that introduces contrastive token-level alignment. 
Specifically, for each speech token, we treat the text token at the same index as a positive pair, while all other tokens serve as negative pairs, excluding identical text tokens at different positions from the negative set.

Table~\ref{tab:ablation_results} (middle) summarizes the results. 
Both variants degrade performance by approximately 15\%, despite higher CKA scores (MSE: 0.6679, InfoNCE: 0.7113). 
In particular, InfoNCE produces a clearer diagonal structure in the token-level similarity maps (Figure~\ref{fig:speech_text_similarity}) due to its contrastive formulation. 
These results suggest that stronger alignment can improve representational similarity without necessarily improving performance, suggesting that overly constraining speech representations toward text may interfere with information that is not preserved in text. We examine this possibility in the following section.

\subsubsection{Why Does Better Alignment Not Always Lead to Better Performance?}

To understand why stronger speech--text alignment does not necessarily improve downstream performance, we analyze linguistic and speech-specific information separately.
We use a text-input variant as a reference for linguistic information, providing ground-truth transcriptions when available and Whisper-large-v3 transcriptions otherwise.
We additionally group the nine AIR-Bench Foundation tasks into linguistic tasks (Speech Grounding, Spoken Language Identification, Speech Entity Recognition, and Intent Classification) and speech-specific tasks based on whether they can primarily be solved from linguistic content.
We exclude Speech-IFEval, whose forgetting-rate metric is defined relative to text-input performance.

First, linguistic understanding alone does not guarantee optimal downstream performance.
As shown in Table~\ref{tab:linguistic_speech_specific}, the text-input variant performs best on the linguistic subset of AIR-Bench Foundation and strongly on AIR-Bench Chat and SpeechR, but substantially worse on the speech-specific subset and MMSU.
This indicates that some speech tasks additionally require acoustic and paralinguistic information absent from text.

Second, stronger internal alignment objectives do not necessarily lead to stable or effective training.
Ours (InfoNCE) achieves higher CKA than our final model but underperforms it across both AIR-Bench Foundation subsets and other downstream benchmarks.
Likewise, removing $\mathcal{L}_{\mathrm{behavior}}$ and relying only on internal alignment causes substantial degradation, suggesting that overly strong alignment can over-constrain speech representations toward text.

Together with Figure~\ref{fig:lambda_ablation} and Section~\ref{sec:Experiments:ablation:internal}, these findings suggest that internal alignment should complement behavior alignment rather than be maximized.

\subsubsection{Effect on Different Query Allocation Strategies}\label{sec:Experiments:ablation:dqa}

We analyze the impact of dynamic query allocation (DQA) by comparing it with a fixed query allocation (FQA) strategy. 
Following prior work \cite{tang2023salmonn}, FQA assigns a constant number of 3 tokens per second of audio, whereas our approach dynamically adjusts the number of queries based on the target token length during training and uses a speech-rate predictor (SRP) at inference.
To isolate the effect of each component, we consider two variants. 
\textit{Ours (FQA)} replaces DQA with FQA during both training and inference, while \textit{Ours (w/o SRP)} retains DQA during training but uses FQA only at inference to simulate prediction errors in SRP. 
Additionally, we compare with a CIF-based variant, \textit{Ours (Cformer)} \cite{wang2024blsp-emo}, where the modality adapter jointly learns to handle both length mismatch and semantic alignment.

Table~\ref{tab:ablation_results} (bottom) summarizes the results. 
\textit{Ours (Cformer)} shows a substantial performance drop, highlighting the benefit of separating length matching from representation alignment.
\textit{Ours (FQA)} also exhibits significant degradation across most benchmarks, whereas \textit{Ours (w/o SRP)} results in only minor performance loss.
These results suggest that the main benefit of DQA comes from length matching during training rather than from precise length prediction at inference.

\subsection{How SLMs Understand Speech-Specific Information}
\label{sec:Experiments:analysis:iemocap}

A potential concern with token-level internal alignment is that close alignment with text may hinder the model's ability to capture speech-specific information.
While MMSU results suggest that the model retains sensitivity to prosodic and paralinguistic cues, we further analyze how such information is reflected internally.

We analyze this on the IEMOCAP~\cite{busso2008iemocap} dataset by prompting the model to classify emotions from three types of inputs: real emotional speech, neutralized cloned speech, and text.
The neutralized speech is generated using Qwen3-TTS \cite{hu2026qwen3} by cloning the real speech, while instructing the model to remove emotional expression.
To examine which parts of the input contribute to the prediction, we compute attention weights from the input to the emotion labels.
Additional details are provided in Appendix~\ref{sec:appendix:iemocap}.

Figure~\ref{fig:analysis_iemocap} presents the results.
For real speech, strong activations appear around regions associated with emotional expression.
When the same utterances are converted into neutralized speech, these activations become noticeably weaker, suggesting that the highlighted regions reflect emotional expression rather than lexical content alone.
In contrast, text inputs show the strongest activations on punctuation tokens.
These findings suggest that, despite token-level internal alignment, the model still captures speech-specific cues absent from transcripts.
Additional examples and audio samples are provided in the supplementary material.

\subsection{Discussion}\label{sec:Experiments:analysis:discussion}

Our results show that speech-text alignment in SLMs should be treated as a balanced objective rather than a quantity to be maximized.
While moderate token-level alignment improves performance, overly strong alignment can increase representational similarity without improving downstream results.
This highlights the need to preserve speech-specific cues while encouraging linguistic correspondence with text.
Although our analysis in Section~\ref{sec:Experiments:analysis:iemocap} suggests that the model retains speech-specific cues, understanding how SLMs encode and balance linguistic content with such cues remains an important direction for future work.

SLMs have also been extended toward broader audio understanding beyond speech.
In such settings, it becomes hard to define a strong correspondence analogous to speech--transcription pairs.
Investigating how Large Audio Language Models (LALMs) can learn and align representations for non-speech audio represents another promising direction for future research.

\section{Conclusion}\label{sec:conclusion}

In this work, we investigate how current Spoken Language Models process speech relative to text and find that their internal representations remain weakly aligned, suggesting persistent structural differences between the two modalities.
We propose a simple framework that addresses this issue by decoupling length matching from semantic alignment and encouraging speech-text correspondence.
Experiments across multiple benchmarks show that our approach improves representational alignment while achieving competitive performance against strong baselines.
Our findings highlight the importance of addressing structural differences between speech and text for more effective SLM training.
\clearpage
\section{Limitations}\label{sec:limitations}

As discussed in Section~\ref{sec:Experiments:analysis:discussion}, this work has several limitations.
First, while our analysis shows that explicitly mitigating structural differences between speech and text and modeling fine-grained internal alignment can improve downstream performance, as shown in Section~\ref{sec:Experiments:ablation}, the extent to which internal alignment should be encouraged may depend on the target task, model architecture, and training data.
Thus, our findings should not be interpreted as suggesting that improving internal speech--text alignment alone will necessarily lead to better downstream performance.

Second, although our analysis in Section~\ref{sec:Experiments:analysis:iemocap} suggests that SLMs can retain speech-specific cues, we do not fully characterize how linguistic content and speech-specific information are jointly encoded and balanced inside the model.
A more detailed analysis of this interaction remains an important direction for future work.

Finally, this work focuses on speech--text correspondence, where paired speech and transcription data provide a natural basis for alignment.
Extending the analysis to broader audio understanding settings is less straightforward, since non-speech audio often lacks a direct textual counterpart.
Investigating how Large Audio Language Models can learn and align representations for non-speech audio remains an interesting direction for future research.

\bibliography{reference}

\appendix
\clearpage
\section{Appendix}\label{sec:appendix}

\subsection{Further Details on Dataset and Implementation}\label{sec:appendix:implementation}

Table~\ref{tab:dataset_summary} summarizes the datasets used for training. In addition to speech transcriptions, the datasets include various speech-specific annotations such as emotion, intent, and gender. These attributes are jointly paired with the corresponding transcriptions and are utilized during response generation, as described in Equation~\ref{eq:gen-response}. 

Table~\ref{tab:instruction_set} presents the instruction set $I$ adopted in our training framework. The instruction set is largely based on prior work \citep{yu2024self} but the Speech Translation category is excluded in our setting since it yields too many language pairs.

After response generation, we apply an additional filtering step to remove low-quality responses using an LLM-as-a-judge strategy. Specifically, we employ the same backbone model, Qwen2.5-7B-Instruct \citep{qwen2.5}, to score each instruction–response pair on a 5-point scale as shown in Table~\ref{tab:judge_scale}. Only samples with a score of 3 or higher are utilized for training.

\begin{table}[t]
\centering
\small
\begin{tabular}{>{\centering\arraybackslash}m{0.8cm} m{6cm}}
\toprule
\textbf{Score} & \textbf{Criterion} \\
\midrule
0 & Useless: ignores the instruction or refuses to respond \\
1 & Poor: barely related, incomplete, or unhelpful \\
2 & Weak: partially relevant but lacking clarity or depth \\
3 & Fair: mostly relevant and informative, but limited in quality \\
4 & Good: relevant, coherent, and reasonably informative \\
5 & Excellent: directly follows the instruction and provides clear, useful content \\
\bottomrule
\end{tabular}
\caption{LLM-as-a-judge evaluation scale.}
\label{tab:judge_scale}
\end{table}

Our modality adapter $\psi(\cdot)$ is implemented using a Q-former architecture with a maximum query length of 512. It adopts the same hidden dimensionality as the LLM and consists of 2 transformer layers with 4 attention heads. During training, we freeze the speech encoder, and apply LoRA to the LLM with conservative settings ($r=2$, $\alpha=2$), as we observed that larger LoRA configurations led to excessive deviation in the LLM behavior. Our model contains approximately 350M trainable parameters.

Training is conducted on 8 NVIDIA H100 GPUs, with a per-device batch size of 10 and 30 gradient accumulation steps to stabilize the token-wise internal alignment loss. Our final model is trained with $\lambda=0.1$ for 8K training steps with a learning rate of $5\times10^{-5}$. Optimization is performed using AdamW with $\beta_1 = 0.9$, $\beta_2 = 0.99$, and $\epsilon = 1\mathrm{e}{-08}$.

\subsection{Additional Details for Speech-Specific Information Analysis}
\label{sec:appendix:iemocap}

We use the following instruction for the emotion classification analysis:

``Please select the most appropriate emotion expressed in the following speech.
Choose only from <angry, happy, sad, neutral, frustrated, excited, fear, surprise, disgust>.
Respond with exactly one label only.''

We construct the input prompts using a chat template that includes both a system prompt and the above instruction.
For the purpose of analysis, the ground-truth emotion label is appended to the input sequence, allowing us to examine attention patterns directed toward the label tokens.
For each layer, we average the attention weights across grouped-query attention (GQA) heads and visualize the resulting attention maps for both speech inputs and text transcriptions.

In some emotion categories, the tokenizer splits the emotion label into multiple tokens.
In such cases, we compute the average attention weights across the corresponding label tokens and visualize them as a single emotion label.
In Figure~\ref{fig:analysis_iemocap}, the attention weights at the emotion label positions are set to zero for clarity.

\begin{table}[t]
\centering
\resizebox{\columnwidth}{!}{
\begin{tabular}{lccc}
\toprule
 & LibriSpeech & LibriTTS & IEMOCAP \\
\midrule
L1 distance & 8.198 & 4.073 & 4.071 \\
Pearson Correlation & 0.971 & 0.972 & 0.920 \\
\bottomrule
\end{tabular}
}
\caption{Performance of our speech rate predictor across datasets. We report L1 distance and Pearson correlation with respect to ground-truth token lengths.}
\label{tab:srp_performance}
\end{table}

\begin{table*}[t]
\centering
\resizebox{0.6\textwidth}{!}{
\begin{tabular}{lcccccc}
\toprule
 & Air-bench & SpeechR & MMSU & Speech-IFeval & Rel. $\Delta$ (\%) $\uparrow$ & CKA \\
\midrule
$\lambda=0.0$ & 7.24 & 52.00 & 53.14 & -17.68 & -10.66\% & 0.6044 \\
$\lambda=0.1$ & \textbf{7.85} & 52.91 & \textbf{53.85} & \textbf{-12.18} & \textbf{0.00\%} & 0.6399 \\
$\lambda=0.2$ & 7.78 & \textbf{54.90} & 53.77 & -13.81 & -2.31\% & 0.6601 \\
$\lambda=0.4$ & 7.57 & 52.17 & 53.12 & -17.48 & -9.20\% & 0.6914 \\
$\lambda=0.8$ & 7.64 & 53.15 & 53.62 & -17.66 & -8.44\% & 0.7266 \\
$\lambda=2.0$ & 7.62 & 50.77 & 51.81 & -23.62 & -14.90\% & \textbf{0.744} \\
\bottomrule
\end{tabular}
}
\caption{
Ablation results across different values of $\lambda$.
We report performance on multiple benchmarks along with representation similarity (CKA).
}
\label{tab:lambda_ablation}
\end{table*}

\begin{table*}[t]
\centering
\resizebox{0.6\textwidth}{!}{
\begin{tabular}{lcccccc}
\toprule
 & Air-bench & SpeechR & MMSU & Speech-IFeval & Rel. $\Delta$ (\%) $\uparrow$ \\
\midrule
instructions=1 & 7.59 & 47.33 & 52.45 & -22.94 & -16.20\% \\
instructions=2 & 7.64 & 48.04 & 52.22 & -23.92 & -16.27\% \\
instructions=3 & 6.96 & 48.46 & 52.10 & -27.71 & -20.34\% \\
instructions=4 & 7.53 & 48.66 & 53.77 & -23.08 & -15.09\% \\
instructions=5 & 7.74 & 50.59 & \textbf{53.85} & -21.51 & -12.35\% \\
instructions=6 & \textbf{7.85} & \textbf{52.91} & \textbf{53.85} & \textbf{-12.18} & \textbf{0.00\%} \\
\bottomrule
\end{tabular}
}
\caption{
Ablation results with varying numbers of instructions.
All metrics improve consistently as the number of instructions increases,
with the largest gains observed when $instructions=6$.
Higher is better for all metrics, while Speech-IFeval is better when closer to zero.
}
\label{tab:inst_ablation}
\end{table*}

\begin{table*}[t]
\centering
\begin{tabular}{l r p{8cm}}
\hline
\textbf{Dataset} & \textbf{Hours} & \textbf{Information} \\
\hline
GigaSpeech \citep{chen2021gigaspeech}       &  10{,}044.55   &  
transcription, data source \\

DailyTalk \citep{lee2022dailytalk}          &  21.67         & 
transcription, emotion, action \\

LJSpeech \citep{ljspeech17}                 &  23.92         &  
transcription \\ 

SLURP \citep{bastianelli2020slurp}          &  26.27         & 
transcription, intent, action, scenario\\ 

VCTK \citep{yamagishi2019cstr}              &  43.89         & 
transcription, age, gender, accent, region \\

Libriheavy \citep{kang2023libriheavy}       &  51{,}024.12   & 
transcription \\

LibriTTS \citep{zen2019libritts}            &  585.83        & 
transcription \\

Librispeech \citep{panayotov2015librispeech}&  961.05        & 
transcription \\

People's speech \citep{galvez2021people}    &  6{,}246.09    & 
transcription \\

AniSpeech \citep{shoukanlabs_anispeech}     &  34.79         & 
transcription \\

EMNS \citep{noriy2023emns}                  &  1.91          & 
transcription, emotion, gender, age \\

EmoV-DB \citep{adigwe2018emotional}         &  9.49          & 
transcription, emotion \\

ESD \citep{zhou2021seen}                    &  13.41         &  
transcription, emotion \\

EXPRESSO \citep{nguyen2023expresso}         &  10.18         & 
transcription, emotion \\

Fair-Speech \citep{veliche2024towards}      &  55.55         &  
transcription, gender, age, first language, socioeconomic background, ethnicity \\

FSC \citep{lugosch2019speech}               &  14.72         & 
transcription, action, object, location \\

GLOBE \citep{wang2024globe}                 &  611.99        & 
transcription, accent, age, gender \\

IEMOCAP \citep{busso2008iemocap}            &  12.44         & 
transcription, gender, speaking rate, pitch, relative dB, emotion, emotion intensity \\

JL-Corpus \citep{clapv2_jl_corpus}          &  1.41          & 
transcription, emotion, country \\

L2-ARTIC \citep{zhao2018l2arctic}           &  27.51         &  
transcription \\

MELD \citep{poria2019meld}                  &  8.72          & 
transcription, emotion \\

STOP \citep{tomasello2023stop}              &  116.59        & 
transcription, gender, speaker nativeness, intent \\

\hline
\end{tabular}
\caption{Summary of datasets}
\label{tab:dataset_summary}
\end{table*}

\begin{table*}[t]
\centering
\begin{tabular}{l p{0.8\textwidth}}
\hline
\textbf{Category} & \textbf{Instruction} \\
\hline
\multirow{3}{*}{Content Repetition}
& 1. Repeat the provided speech, ensuring to maintain its original meaning and details. \\
& 2. Provide the speech exactly as given—do not alter wording, structure, or omit any content. \\
& 3. Echo the content of the speech, maintaining its exact purpose and details. \\
\hline
\multirow{3}{*}{Keyword Extraction}
& 1. Extract the most frequently occurring words or phrases in the speech, excluding common stopwords, to identify main topics. \\
& 2. Identify and list the most common words or phrases from the speech, omitting typical stopwords, to highlight central themes. \\
& 3. Extract significant words or phrases that appear often in the speech, exclude basic stopwords, to uncover the main subjects. \\
\hline
\multirow{3}{*}{Intent Recognition}
& 1. Determine the primary purpose of the speech and evaluate how clearly and effectively the message is conveyed. \\
& 2. Identify the main intent of the speech and assess the clarity and effectiveness of its delivery. \\
& 3. Assess the central purpose of the speech and evaluate the directness and impact of its expression. \\
\hline
\multirow{3}{*}{Sentiment Analysis}
& 1. Determine the sentiment of the speech and identify which sections contribute most to sentiment. \\
& 2. Evaluate the emotional tone of the speech and determine which segments primarily affect the sentiment. \\
& 3. Assess the sentiment expressed in the speech and highlight which areas contribute most to this feeling. \\
\hline
\multirow{3}{*}{Continuation}
& 1. Please write a coherent and engaging continuation of the given speech with less than 50 words. \\
& 2. Compose a logical and captivating follow-up to the provided speech within 50 words. \\
& 3. Write a fluent and engaging continuation of the speech, limited to 50 words. \\
\hline
\multirow{3}{*}{Speech Recognition}
& 1. Provide the transcription according to the speech. \\
& 2. Convert the spoken language into a written transcript. \\
& 3. Write down the speech as a text transcript. \\
\hline
\end{tabular}
\caption{Instruction Set}
\label{tab:instruction_set}
\end{table*}

\subsection{Artifact Licenses}
\label{sec:appendix:licenses}
We use publicly available datasets (Table~\ref{tab:dataset_summary}), benchmarks, and pretrained models in accordance with their respective licenses and terms of use. 
The benchmarks used in our experiments, including Air-Bench, SpeechR, MMSU, and Speech-IFeval, are used solely for research evaluation. 
We do not redistribute the original datasets, benchmark data, or model checkpoints. 
For pretrained models and codebases, we follow the licenses and usage conditions specified by the original authors or providers.

\end{document}